\documentclass[accepted]{uai2024} 
                        
\usepackage[american]{babel}
\usepackage{comment} 

\usepackage{natbib} 
\usepackage{mathtools} 
\usepackage{booktabs} 
\usepackage{tikz} 
\usepackage{amssymb} 
\usepackage{dsfont}

\title{MetaCOG: A Hierarchical Probabilistic Model for Learning Meta-Cognitive Visual Representations}

\author[1]{\href{mailto:<marlene.berke@yale.edu>?Subject=Your UAI 2024 paper}{\vspace{-0.5em}Marlene~D.~Berke~}{}}
\author[2]{Zhangir~Azerbayev}
\author[1]{Mario~Belledonne}
\author[3,4]{Zenna~Tavares}
\author[1,5]{Julian~Jara-Ettinger}
\affil[1]{%
    Psychology Dept.\\
    Yale University\\
    New Haven, Connecticut, USA
}
\affil[2]{%
    Computer Science Dept.\\
    Princeton University\\
    Princeton, New Jersey, USA
}
\affil[3]{%
    Basis Research Institute\\
    New York, New York, USA
  }
\affil[4]{%
    Zuckerman Institute and Data Science Institute\\
    Columbia University\\
    New York, New York, USA
  }

\affil[5]{%
    Wu Tsai Institute\\
    Yale University\\
    New Haven, Connecticut, USA
  }
  
\begin{document}
\maketitle

\begin{abstract}
Humans have the capacity to question what we see and to recognize when our vision is unreliable (e.g., when we realize that we are experiencing a visual illusion). Inspired by this capacity, we present MetaCOG: a hierarchical probabilistic model that can be attached to a neural object detector to monitor its outputs and determine their reliability. MetaCOG achieves this by learning a probabilistic model of the object detector's performance via Bayesian inference---i.e., a \textit{meta-cognitive} representation of the network's propensity to hallucinate or miss different object categories. Given a set of video frames processed by an object detector, MetaCOG performs joint inference over the underlying 3D scene and the detector's performance, grounding inference on a basic assumption of object permanence. Paired with three neural object detectors, we show that MetaCOG accurately recovers each detector's performance parameters and improves the overall system's accuracy. We additionally show that MetaCOG is robust to varying levels of error in object detector outputs, showing proof-of-concept for a novel approach to the problem of detecting and correcting errors in vision systems when ground-truth is not available.
\end{abstract}

\section{Introduction}\label{sec:intro}

Building accurate representations of the world is critical for prediction, inference, and planning in complex environments \citep{lake2017building}. While the last decade has witnessed a revolution in the performance of end-to-end object recognition models, these models can nonetheless suffer from errors, particularly when confronted with out-of-sample data. How can we identify when a detected object is not actually there (a hallucination), or when an object present in a scene was not detected (a miss)?


Traditional approaches to this problem involve training the model to express intrinsic uncertainty over its outputs (e.g., \citet{sensoy2018evidential}; \citet{kaplan2018uncertainty}; \citet{ivanovska2015subjective}). While powerful, model performance can change when deployed in contexts not well represented in training. And sometimes, appropriate datasets for retraining are not always readily available. Here we focus on this class of problems. On these problems, the approach of post-hoc uncertainty learning (which seeks to extrinsically quantify uncertainty over a model's outputs) has the potential improve the reliability of a pre-trained off-the-shelf model (e.g., as in \citet{shen2023post}).

This paper draws inspiration from human cognition to introduce a proof-of-concept solution to the problem of post-hoc uncertainty learning. While human vision is generally robust and reliable, it nonetheless suffers from occasional errors, such as when we experience a visual illusion, like a mirage on a highway or apparent motion in a static image. Even though these illusions fool our visual system (i.e., they look real), people recognize that they shouldn't be trusted. This is because humans have internal models of how their own vision works that help them decide when to trust or mistrust what they see \citep{berke2024core}---a form of \textit{meta-cognition} \citep{nelson1990metamemory}. We propose that a similar approach can be used to improve the robustness of object detection models, by combining object detectors with an external model that represents the object detector's behavior (Fig. \ref{Fig:conceptual}) and evaluates its outputs, integrating uncertainty and flagging unreliable detections. We present a formalization of this idea and a proof-of-concept of the approach.

There are two main sets of observations and insights that we draw upon. First, human vision is an encapsulated black-box system and we do not have access to its internal computations \citep{firestone2016cognition}. Yet, people are still able to learn a model of their own visual system \citep{berke2024core}. This suggests that such learning is possible. We therefore focus on a problem formulation where the meta-cognitive system does not have access to the internal architecture of the object detector---only its outputs. Second, people can detect failures in their visual system spontaneously and without need for feedback \citep{berke2024core} and we therefore aim to do the same in our model. Our insight is that humans embed object detections in three-dimensional representations that assume object permanence (i.e., objects in the world continue to exist even when they cannot be seen; \citet{carey2009origin}; \citet{spelke2007core}). For instance, if you detect an object in your periphery but it disappears when you look directly at it, you can conclude your visual system made an error, because objects do not spontaneously cease to exist.

Inspired by this, our model, MetaCOG (Meta-Cognitive Object-detecting Generative-model), aims to monitor an object detector's output and identify which detections were hallucinated and which objects were missed, without feedback or access to ground-truth object labels. MetaCOG achieves this by learning a model of the neural object detector's performance (i.e., a visual meta-cognition), represented as a joint distribution between objects in a scene and outputs produced by the object detector. Specifically, this probability distribution represents the object detector's propensity to hallucinate or miss objects of each visual category.

We evaluate MetaCOG's ability to 1) learn an accurate representation of the object detector's performance and 2) use this meta-representation to identify and correct missed or hallucinated objects. In Exp. 1, we explore MetaCOG's performance when paired with three modern neural object detectors, testing MetaCOG on a dataset of scenes rendered in the ThreeDWorld (TDW) virtual environment \citep{gan2020threedworld}. In Exp. 2, we explore MetaCOG's robustness by systematically testing its tolerance to different degrees of faulty inputs.

Our contributions are as follows: First, we propose a novel model, MetaCOG, that implements a human-inspired method for improving the performance of object detectors. Second, we introduce a new representation and inference method that enables accurate recovery of an object detector's propensity to miss and hallucinate object categories, without feedback or access to ground truth. Third, we show that, as a byproduct, MetaCOG's detection of misses and hallucinations allows us to automatically generate a tailored dataset of corrected errors, which may be used to fine-tune the object detector and improve its performance. Finally, we show through synthetic experiments that our approach is successful for a wide range of object detector performances.

\section{Related Work}

\paragraph{Meta-cognition in AI.} Previous work has shown the promise of meta-cognition for improving classification accuracy \citep{babu2012sequential,subramanian2013metacognitive}. While that work focused on engineering (rather than learning) a meta-cognition to guide training, we focus on a complementary problem: learning a meta-cognition for correcting outputs from a pre-trained network.

\paragraph{Object knowledge.} Our work is also related to computational models of infant object knowledge \citep{smith2019modeling,kemp2009ideal} and work applying cognitively-inspired object principles to computer vision \citep{chen2022unsupervised,tokmakov2021learning}. The difference is that MetaCOG uses object principles to learn a meta-cognition, whereas past work focused on modeling the object principles themselves.

\paragraph{Neurosymbolic AI.} Our work can be seen as an instance of neurosymbolic AI \citep{garcez2020neurosymbolic}. Related work used detections from neural object detectors and a generative model of scenes to infer a symbolic scene graph \citep{gothoskar20213dp3}, but lacked the meta-cognitive component central to our work. In the language domain, \citet{nye2021improving} used a symbolic world model to improve the coherence of a large language model, similar to how MetaCOG's symbolic world models and meta-cognition improve the coherence of object detections. Our work is unique in its formulation of meta-cognition within a neurosymbolic framework.


\section{MetaCOG}

\begin{figure*}[t]
\centering
\includegraphics[width=\linewidth]{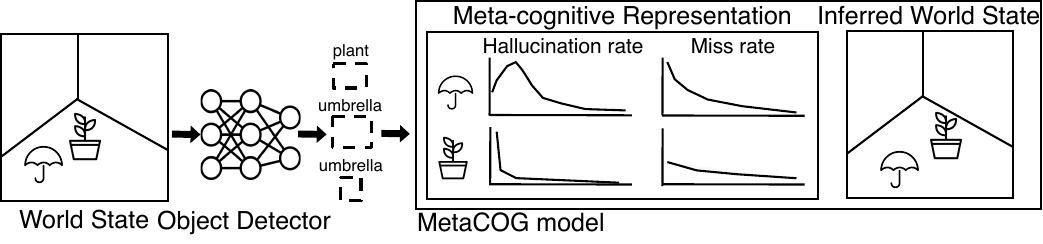}
\caption{Conceptual schematic of MetaCOG. Information flows from left to right. From the left, images of a scene (described by a ``World State") are processed by a neural ``Object Detector," which produces detections (semantic labels with bounding boxes in 2D images). The MetaCOG model takes detections as input (without any access to the underlying world state or to the ground-truth accuracy of these detections), and jointly infers a meta-cognitive representation of the detector and what objects are where in the scene (``Inferred World State"). Specifically, MetaCOG's ``Meta-cognitive Representation" consists of beliefs about the category-specific probabilities of the object detector generating hallucinations (detections of objects that were not actually there) and misses (failures to detect objects that were actually there). MetaCOG simultaneously infers this meta-cognitive representation and the world state (i.e., semantic labels and locations in 3D space).}\label{Fig:conceptual}
\end{figure*}


Throughout, we consider the following problem formulation. Given a physical space (e.g., a room with objects in it), a set of images is sampled from a known camera trajectory $\vec{cj}$ (such that objects may pass in and out of view) and processed by an object detector. The goal of MetaCOG is to use this series of outputs from the object detector to learn about the detector's performance (i.e., what categories the detector is likely to miss or hallucinate), and to infer the underlying state of the physical space (i.e., the semantic label and 3D position of each object in the scene).

Fig. \ref{Fig:conceptual} illustrates MetaCOG's pipeline. Given a world state that is passed through an object detector, a list of object detections (which include their 2D location) is provided as input to MetaCOG. MetaCOG then uses Bayesian inference to jointly infer (1) a meta-cognitive representation and (2) the underlying world state (i.e., objects and their location in 3D space). The meta-cognitive representation is formed by two category-specific probability distributions: one capturing the probability of hallucinated detections (detected, but not actually there) for each category, and another capturing the probability of missed detections (not detected, but actually there) for each category. This representation is meta-cognitive in the sense that its content references not the external world, but the vision system itself. Due to its Bayesian nature, MetaCOG produces a posterior distribution over world states, and a posterior distribution over meta-cognitive representations. After repeating this process over multiple different world states, we can estimate the object detector's true probabilities of hallucinating and missing objects by taking the expectation of the posterior over meta-cognitive representations. The world states can be estimated using the MAP of the posterior over world states. 

The rest of this section describes MetaCOG more formally, starting with how world states and detections are represented, building up to the meta-cognitive representation and the kernel for updating it, and finally concluding with the inference procedure.

\subsection{Generative model}\label{sc:generativemodel}

\paragraph{World states and object constraints}

A world state is represented as a collection of objects, each object a 4-tuple of the form $(x, y, z, c)$ where $x$, $y$, and $z$ are coordinates in 3D space and $c$ is an object category. We denote by $\vec{W} = (W_1, \dots, W_T)$ a sequence of world states.

Our approach implements object constraints into world states, which we hypothesize will make the visual meta-cognition learnable without access to ground-truth objects. Inspired by human infants' representations of objects, we assume that objects occupy locations in 3D space and that no two objects can occupy overlapping positions (implemented as a prior over the locations of the objects in a world state; details available in \ref{Additional generative model parameters}). In this work, objects were assumed to be stationary, but the approach can extend to moving objects.\footnote{The approach could be extended to moving objects by leveraging the human-inspired expectation that objects have spatio-temporal continuity, and therefore move in smooth and continuous trajectories \citep{spelke2007core}. This could be implemented as a prior over object motion. Rather than inferring a single location for each object, the problem becomes inferring a trajectory for each object.} MetaCOG implements object permanence as the constraint that each world state has a fixed set of objects that does not change (as a consequence, an object in a scene is still there even when it is not observed due to being out of view, occluded, or missed by the detector).

In our setting, multiple images are taken of each scene in a sequence of world states $\vec{W}$. We assume access to the camera's position and orientation for each image, which also implies knowledge of a scene change. While inferring the camera trajectory is possible \citep{taketomi2017visual}, that is not the focus of our work. Critically, the MetaCOG model does not have access to the ground-truth objects in the underlying world states, and must instead infer them from detections output by a faulty object detector.

\paragraph{Object detector outputs and inputs to MetaCOG}

Given an image, a detector produces an unordered list of detections, each consisting of a category label and a position on the 2D image. Each detection is a tuple $(x, y, c)$ where $x$ and $y$ are pixel-coordinates (i.e., the centroid of a bounding box) and $c \in \mathbb{C}$ is an object category. The neural network object detector can be seen as a function $NN$ applied to an image $i$, producing a collection of detections, call this $NN(i)$. If $\mathbb{I}_{W_t}$ is the collection of images taken of the scene described by world state $W_t$, then the collection of detections generated from all of those images is $NN(\mathbb{I}_{W_t})$. For convenience, we will refer to this collection of detections generated from the world state $W_t$ as $D_t$. Crucially, while MetaCOG has access to the detections generated by a detector, the object detector itself is a black box---MetaCOG does not have any access to its internal state.

\paragraph{Meta-cognitive representation}\label{meta-cognitive dynamics}

The meta-cognition represents two aspects of the object detector's performance: its propensity to hallucinate objects that are not there, and its propensity to miss  (or conversely, accurately detect) objects that are there. This is represented as two probability distributions per object category. The first distribution captures the detector's propensity to hallucinate, modeled as the number of times objects of category $c \in \mathbb{C}$ will be hallucinated in a given frame. Under naive assumptions, hallucinations may be independent of each other and randomly distributed within and across images, therefore following a Poisson distribution with rate $\lambda_c$ (to be inferred by MetaCOG). The second distribution captures the detector's propensity to correctly detect an object that is in view. Because object detectors can produce multiple detections from a single object (e.g., if it is incorrectly parsed as two objects), this representation follows a Geometric distribution with rate $p_c$ encoding a belief over the number of times ($0, 1, ..., k$) an object of category $c$ will be detected when it is present in the image. Conveniently, under this formulation, the detector's miss rate for an object in category $c$ is $1-p_c$.

Because each distribution is captured by a single parameter, the parameters of the distributions forming the meta-cognition can be represented as a pair of vectors of length $|\mathbb{C}|$ storing each category's hallucination rate $\lambda_c$ and miss rate $1-p_c$. We call this pair of vectors of parameters $\theta$. We refer collectively to the distributions that they parameterize as $V$ since they express a belief about the veracity of the object detector. For notation simplicity, $V$ will refer to $V|\theta$.

The model described so far captures how MetaCOG represents an object detector's performance. However, a detector's propensity to hallucinate or miss objects can vary across scenes, so leaving flexibility in $V$ is desirable. At the same time, experience in a previous scene includes critical information about the detector that should inform expectations about its performance in a new scene. Our generative model therefore includes an evolving kernel, capturing changing beliefs over the parameters $\theta$ for the probability distributions $V$. 
Since beliefs about a detector evolve from scene to scene, $V_t$ denotes the belief at time $t$, and $\vec{V} = (V_1, ..., V_T)$ denotes a sequence of evolving beliefs over timesteps.

\paragraph{Meta-cognitive learning kernel}
The priors over the hallucination and miss rates are updated by matching up the objects that MetaCOG infers to be present in a scene with the detections that they caused. This allows MetaCOG to identify which objects were missed, and which detections were hallucinated. The details of the procedure for matching up objects and their detections are given in \ref{diff between Wt and Ot}).

Since the Gamma distribution is the conjugate prior of the Poisson distribution representing hallucinations, the Gamma distribution is used to model beliefs about each category's hallucination rate $\lambda_c$. At $t=0$, the parameters of each Gamma prior are initialized to $\alpha = 1$ and $\beta = 1$ (representing a relatively uninformative prior with a mean of 1). After inference over a scene, the Gamma prior over $\lambda_c$ evolves by computing the inferred number of hallucinations at time $t$ (based on the difference between world state $W_t$ and associated detections $D_t$, see \ref{diff between Wt and Ot}), and updating the $\alpha$ and $\beta$ parameters of the Gamma distribution.

Beliefs about the detection rates $p_c$ evolve in an analogous manner. As Beta is the conjugate prior of the Geometric distribution representing detections, beliefs about the miss rate evolve by updating the $\alpha$ and $\beta$ parameters of the Beta distribution. At $t=0$,  the Beta prior is initialized with parameters $\alpha = 1$ and $\beta = 1$ (equivalent to a uniform distribution on $[0, 1]$, a maximally uninformative prior for detection rate). The parameters from the Beta distribution are then updated based on the inferred missed detections (by comparing $W_t$ against $D_t$, see \ref{diff between Wt and Ot}).

\begin{figure}[h]
\centering
\includegraphics[width=0.9\linewidth]{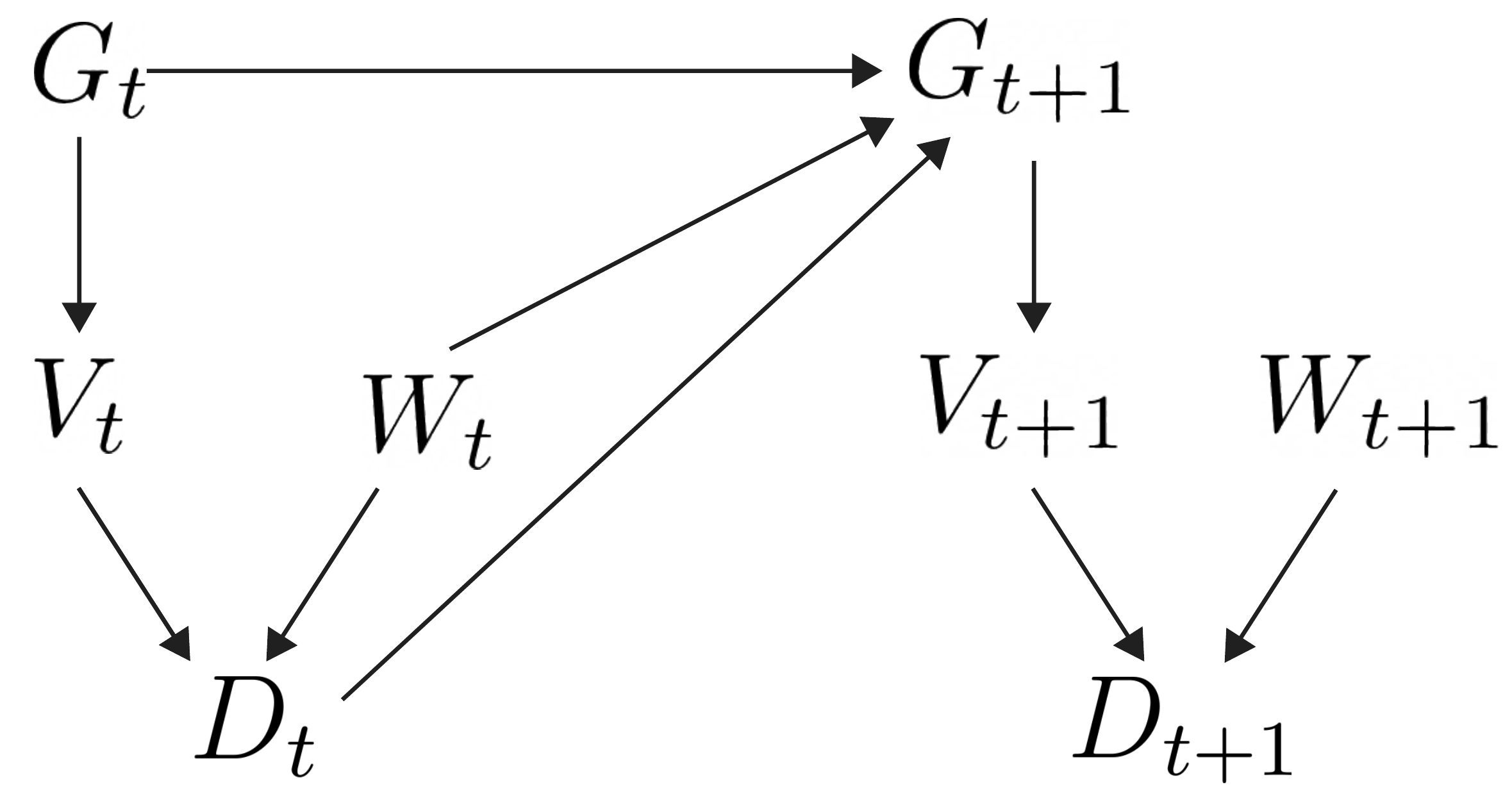}
\caption{Schematic depicting the forward generative model of how world states ($W$) and a vision system with some veracity ($V$) produce detections ($D$). $G_t$ is the prior over the meta-cognitive representation ($V_t$) at time $t$; $W_t$ is the world state; and $D_t$ are the detections that are generated. $W_t$, $D_t$, and $G_t$ collectively influence $G_{t+1}$, the prior over $V_{t+1}$.}
\end{figure}

\subsection{Inference procedure}\label{sc:inference}

So far, we have described MetaCOG as a model of the generative process where 2D images of scenes are processed by an object detector to produce collections of detections. These detections, along with camera trajectories, serve as observations used to infer world states and a meta-cognitive representation of the object detector.


Given $\vec{D}$ (collections of detections ($D_1, ..., D_T$) from a sequence of world states $\vec{W}$) and the corresponding camera trajectories $\vec{cj}$, the goal is to jointly infer (via Bayesian inference) the evolving meta-cognitive representation $\vec{V}$ and the world states $\vec{W}$ underlying each scene:

\small
\begin{equation*}
P \left(\vec{V},\vec{W} \mid \vec{D},\vec{cj} \right) \propto \\         \prod_{t=1}^{T}P\left(D_t|W_t, V_t,cj_t\right)P(V_t)P(W_t)
\end{equation*}
\normalsize
\label{vr_joint_posterior}The posterior is approximated via Sequential Monte-Carlo using a particle filter (for background, see \citeauthor{doucet2009tutorial} \citeyear{doucet2009tutorial}). The details of the algorithms used for inference are described in \ref{Inference Procedure Details}.

An estimate of the joint posterior can be sequentially approximated via:

\small
\begin{equation*} \label{eq:vr_seq_estimate}
\begin{split}
    & P\left(\vec{V}, \vec{W} | \vec{D}, \vec{cj}\right) \approx P\left(\hat{V_0}^0\right) * \\[-5pt] 
    & \prod_{t=1}^{T}
    P\left(D_t | \hat{V_t}^t, \hat{W_t}^{t}, cj_t\right)
    P\left(\hat{W_t}^t\right) P\left({\hat{V_t}}^t | {\hat{W_{t-1}}^t}, {\hat{D_{t-1}}^t}\right)
\end{split}
\end{equation*}
\normalsize
where $\hat{W_1}^T, \ldots, \hat{W_T}^T$ is the estimate of $W_1, \ldots, W_T$ after detections from $T$ world states have been observed, and $\hat{V_1}^T, \ldots, \hat{V_T}^T$ is the estimate of $V_1, \ldots, V_T$ after $T$ observations. 
Here the transition kernel, $Pr(\hat{V_t}^t | \hat{{W_{t-1}}^t}, {\hat{D_{t-1}}^t})$ is governed by the meta-cognitive learning kernel. 

\paragraph{Estimating $V$}\label{Estimating V}

After all $T$ world states have been processed, we estimate $V_T$ by taking the expectation of the marginal posterior distribution by averaging across particles weighted by their likelihood $l$: $\hat{V}_{T,\mu}^T = E[\hat{V}_T^T | \vec{D}] = \frac{1}{M} \sum_{m=1}^{M}(\hat{V}_{T,m}^T * l_m)$, where $m$ indexes the particles. This $\hat{V}_{T,\mu}^T$ is the final estimate of the belief about the true $V$ after all detections have been observed. Given $\hat{V}_{T,\mu}^T$, the conditional posterior can be estimated as:

\small
\begin{equation*} \label{eq:post_pred_dist}
\begin{split}
& P\left(\hat{\vec{W}} | \vec{D}, \vec{cj}, \vec{V} = \hat{V}_{T,\mu}^T\right) \propto \\[-5pt] 
& \prod_{t=1}^{T}P\left(D_t|\hat{W}_t, cj_t, V_t = \hat{V}_{T,\mu}^T\right)P\left(\hat{W}_t\right)
\end{split}
\end{equation*}
\normalsize
This conditional posterior can be used to infer world states for novel scenes, $W_{T+1},...$ or to reassess previous world states $W_1,... ,W_T$ that were originally inferred using a less informed meta-cognitive representation $\hat{V}$.

\section{Experiments}

Our experiments have two goals: first, to test whether MetaCOG can learn an accurate meta-cognitive representation of an object detector, and second, to explore whether this meta-cognitive representation confers any benefits in overall accuracy or robustness to faulty inputs. Exp. 1 tests MetaCOG's performance when processing outputs of three popular object detection systems (Section \ref{sc:exp1}). Exp. 2 presents a robustness analysis of how MetaCOG performs as a function of an object detector's baseline performance (Section \ref{sc:exp2}). Throughout, we evaluate MetaCOG by sampling scenes (arrangements of objects in a room), and then sampling images taken from different viewpoints from each scene (calling the collection of images from a scene a ``video"). Model code, datasets, results, and demos are available: \url{https://github.com/marleneberke/MetaCOG/tree/main}.

\begin{figure*}[t]
\centering\includegraphics[width=1.0\linewidth]{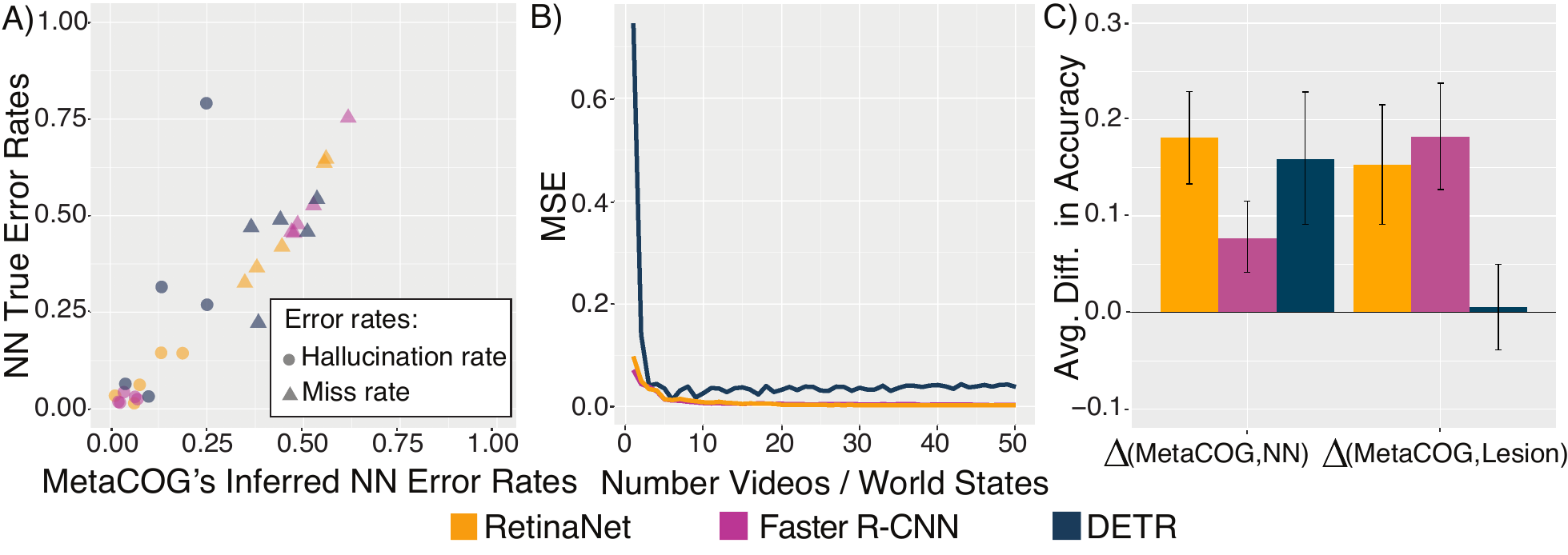}
\caption{Results for MetaCOG and comparison models. Throughout, yellow codes for RetinaNet, magenta for Faster R-CNN, and blue-grey codes for DETR. A) Scatterplot showing MetaCOG's inferred values for the hallucination rates (circles) and miss rates (triangles) against the ground-truth values. B) The MSE (averaged across categories) of MetaCOG's inferences about $\theta$ as a function of number of videos observed. C) Comparisons between MetaCOG and the two baseline models on the test set (after conditioning on the meta-cognitive representation that MetaCOG inferred on the training set). The left group of bars show the difference between MetaCOG and the NN's output, and the right shows the difference between  MetaCOG and the lesioned model. Positive values indicate MetaCOG outperforms the comparison model.}\label{Fig:Exp1Results}
\end{figure*}

\subsection{Experiment 1: Enhancing neural networks for object detection with a meta-cognition}\label{sc:exp1}

\begin{table*}[h]
\centering
\begin{small}
\begin{sc}
\caption{Accuracy Results for Exp. 1. Values in Parentheses Are 95\% Bootstrapped Confidence Intervals.}
\label{MainTable}
\begin{tabular}{llcc}
\toprule
Object Detector & Model            & Acc. Training                   & Acc. Test\\
\midrule
RetinaNet       & MetaCOG          & \textbf{0.72} (0.66, 0.79)   & \textbf{0.72} (0.65, 0.77) \\
                & NN Output        & 0.56 (0.52, 0.61)            & 0.54 (0.50, 0.57) \\
                & Lesioned MetaCOG & 0.64 (0.56, 0.71)            & 0.56 (0.50, 0.63) \\
Faster R-CNN     & MetaCOG          & \textbf{0.79} (0.73, 0.84)   & \textbf{0.76} (0.71, 0.82)\\
                & NN Output        & 0.75 (0.70, 0.80)            & 0.69 (0.64, 0.74)\\
                & Lesioned MetaCOG & 0.66 (0.59, 0.73)            & 0.58 (0.52, 0.64)\\
DETR            & MetaCOG          & 0.32 (0.25, 0.39)            & \textbf{0.32} (0.24, 0.39) \\
                & NN Output        & 0.16 (0.13, 0.18)            & 0.16 (0.13, 0.18) \\
                & Lesioned MetaCOG & \textbf{0.36} (0.29, 0.44)   & 0.31 (0.24, 0.38) \\
\bottomrule
\end{tabular}
\end{sc}
\end{small}
\end{table*}

\paragraph{Object detection models}\label{objectmodels} 

To test MetaCOG's capacity to learn and use a meta-cognitive representation, we tested its performance when paired with three modern neural networks for object detection. These networks represent three popular architectures: RetinaNet, a one-stage detector (\citeauthor{lin2017} \citeyear{lin2017}); Faster R-CNN, a two-stage detector (\citeauthor{ren2015} \citeyear{ren2015}); and DETR, a transformer (\citeauthor{carion2020} \citeyear{carion2020}), all pre-trained. The networks were validated to ensure that their baseline performances on our dataset were within their expected ranges. See \ref{post-processing} for details.

\paragraph{Dataset}\label{Dataset Exp 1}
We evaluate MetaCOG on a dataset rendered in the ThreeDWorld physical simulation platform (\citeauthor{gan2020threedworld} \citeyear{gan2020threedworld}) using one canonical object model per category. First, 100 scenes were generated by sampling objects and placing them in a room with carpeting and windows. For each scene, we then generated a video by sampling 20 frames from the ego-centric perspective of an agent moving around the room. The resulting set of 100 videos was then randomly split into a ``training"\footnote{Note that the training set is not used for training in the traditional sense, since the ground-truth object labels are never used.} and test set (each with n=50 videos). To avoid order effects, all reported results show averages across four different video orders. See \ref{Experiment 1 Dataset} for the motivation for this dataset, sample images, and further details.

\paragraph{Comparison models}\label{baselines}
We compare MetaCOG to two baselines in order to test if learning a meta-cognitive representation improves performance. First, we compare MetaCOG's inferences about world states to the neural network's post-processed detections serving as input to MetaCOG (see \ref{post-processing} for details). Second, it is possible that the computational structure of MetaCOG might improve accuracy without meta-cognition per se mattering. For instance, mapping detections to 3D representations with object permanence alone might improve accuracy. Alternatively, having a meta-cognitive representation might provide tolerance for suppressing hallucinations or recovering missed objects, but the ability to infer it (and the resulting values of the parameters $\theta$ in $V$) might not matter. To test these possibilities, we compare our results against a \textit{Lesioned MetaCOG}, where the values of $\theta$ were set to the mean of the initial prior over $\theta$ (see \ref{lesion}). Although this model has a meta-cognitive representation $V$, the values of its parameters $\theta$ are neither learned nor updated in light of observations. Since this lesioned model contains the same representations as the full model, it serves as a control for model complexity.

\paragraph{Results}

We first assessed MetaCOG's ability to learn an accurate meta-cognitive representation $V$ (see \ref{MSE for V} for details). Fig. \ref{Fig:Exp1Results}A shows the relationship between the parameters in MetaCOG's learned $\hat{\theta}$ and each network's true $\theta$, with an overall correlation of $r=0.878$. Fig. \ref{Fig:Exp1Results}B shows the learning trajectory, visualizing the MSE of $\hat{\theta}$ for each network as a function of experience (observed detections from videos). After 50 videos, MSEs decreased, on average, by 96.1\% of their initial values, and had already decreased by 94.9\% after only 20 videos. This demonstrates that MetaCOG can learn an accurate meta-cognition $V|\theta$ efficiently and without access to the ground-truth object labels.

Table \ref{MainTable} and Fig. \ref{Fig:Exp1Results}C show MetaCOG's accuracy relative to comparison models (see \ref{Metrics for Exp 2} for operationalization). MetaCOG outperformed the outputs of all three neural networks, and for two of the three, MetaCOG also outperformed the lesioned model. For DETR, we did not see a significant difference in accuracy between MetaCOG and the lesioned version, perhaps because Lesioned MetaCOG was already performing well due to a coincidence where the priors over $\theta$ gave an adequate approximation of DETR's performance on this dataset (see \ref{failure} for further discussion). When paired with Faster R-CNN, the full MetaCOG model improved the system's overall accuracy, but the lesioned model actually impaired performance. This is because the meta-cognitive representation in the lesioned model was a very poor description of Faster R-CNN's behavior. In particular, Faster R-CNN had very low hallucination rates ($\lambda_c < 0.05$ for every object category), while the lesioned model fixed all $\lambda_c$s to the mean of their priors ($\lambda_c=1.0$). Because lesioned MetaCOG incorrectly represented Faster R-CNN as tending to hallucinate much more than it actually did, lesioned MetaCOG was overly skeptical of Faster R-CNN's detections and incorrectly dismissed many correct detections. This actually impeded overall performance, demonstrating the importance of MetaCOG’s ability to learn and update its meta-cognitive representation as a function of experience. Sometimes, a poor or misleading meta-cognition may be worse than no meta-cognition at all.

Overall, averaged across the three NNs, MetaCOG increased accuracy by $13.9\%$ relative to the NN outputs in the test set. On average, MetaCOG also increased accuracy by $11.4\%$ relative to Lesioned MetaCOG, confirming that MetaCOG's success can be partially attributed to its learned meta-cognition, rather than purely to the 3D object representation and priors (see \ref{3D results} for additional results showing that MetaCOG's inferences about the presence and locations of objects at the level of 3D scenes rather than 2D images also outperform those of Lesioned MetaCOG. Also see \ref{Additional dataset} for results on an additional dataset).

While MetaCOG and Lesioned MetaCOG were matched in architecture, computational complexity, and input data, MetaCOG and NN Output were not. This raises the possibility that MetaCOG's success relative to the neural network's outputs could be due to its additional structure, computational complexity, and access to camera trajectories. To create a like-to-like comparison, we fine-tuned Faster R-CNN using MetaCOG's inferences on the training set (see \ref{Training Faster R-CNN} for details) and compared it to off-the-shelf Faster R-CNN. It is possible to use MetaCOG to fine-tune an object detector since MetaCOG's inferences about what objects are where in 3D space can be projected back into 2D space to label the images, creating synthetic, labeled training data (all without access to ground-truth object labels). We compared Faster R-CNN's performance on the test set, before and after fine-tuning.

\begin{figure}[t]
\centering
\includegraphics[width=0.8\linewidth]{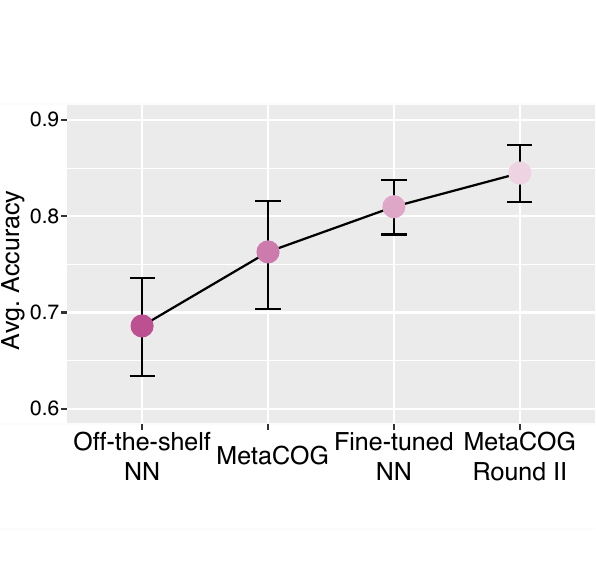}
\caption{The results of MetaCOG fine-tuning Faster R-CNN. Each point shows the average accuracy of a model on the test set, and the bars are bootstrapped 95\% CIs. From left to right, the leftmost point shows the accuracy of pre-trained, off-the-shelf Faster R-CNN. The next point shows the accuracy of MetaCOG, when paired with Faster R-CNN's inputs. The difference between these two points is depicted by the magenta $\Delta$(MetaCOG, NN) bar in Fig. \ref{Fig:Exp1Results}C. The third point shows the accuracy of Faster R-CNN after fine-tuning using MetaCOG's inferences. The rightmost point shows the accuracy of MetaCOG with inputs from fine-tuned Faster R-CNN. For exact accuracy values, see Table \ref{retraining_table} in \ref{Training Faster R-CNN}.}
\label{Fig:retraining}
\end{figure}

Although MetaCOG's inferences had an accuracy of only 0.76 (Fig. \ref{Fig:retraining} MetaCOG), fine-tuning Faster R-CNN using MetaCOG's inferences improved the network's accuracy on the test set from 0.69 (Fig. \ref{Fig:retraining} Off-the-shelf NN) to 0.81 (Fig. \ref{Fig:retraining} Fine-tuned NN). Comparing Faster R-CNN before and after fine-tuning allows for a like-to-like comparison showing the impact of MetaCOG while controlling for computational complexity. Furthermore, these results serve as a proof-of-concept that MetaCOG can be used to train an object detection system. 

Paired with this fine-tuned object detector, can MetaCOG use the NN's improved detections to make even better inferences? With more accurate inputs (0.81) MetaCOG did indeed draw even more accurate inferences (0.85; see MetaCOG Round II in Fig. \ref{Fig:retraining}). These results suggest that iteratively improving the object detector via fine-tuning and MetaCOG via better inputs has the potential to create a self-supervised learning system.

\subsection{Experiment 2: Robustness analysis}\label{sc:exp2}

\begin{figure*}[t]
\centering\includegraphics[width=\linewidth]{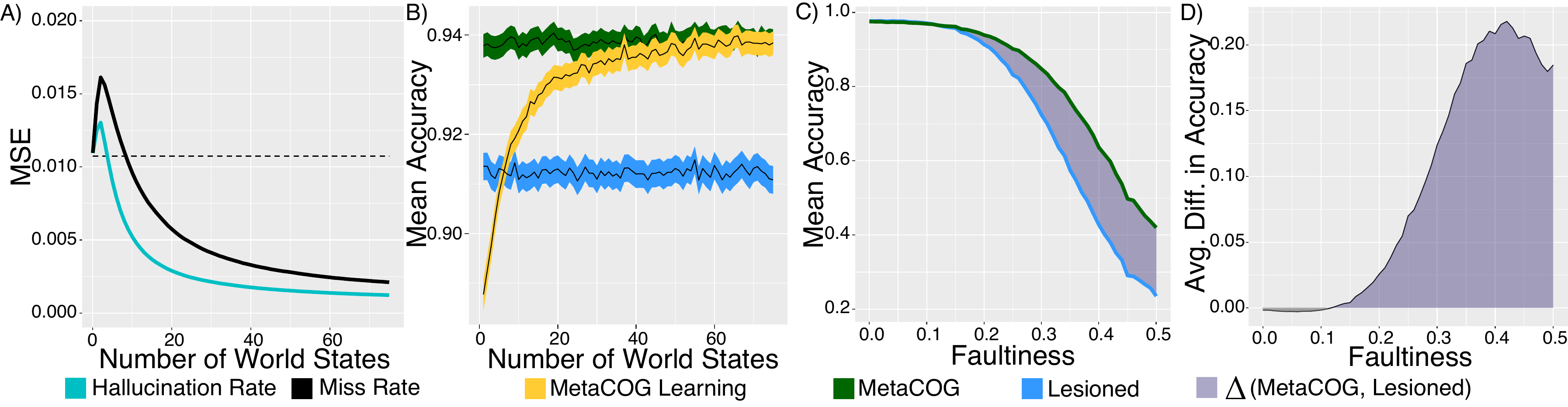}
\caption{Lightweight MetaCOG's average performance over 40000 simulated detectors varying in faultiness, each processing 75 world states. \textbf{A)} MSE between true and inferred hallucination and miss rates as a function of number of processed world states. Horizontal dotted line represents the average MSE for the mean of the prior (the $\theta$ used in Lesioned MetaCOG). \textbf{B)} Mean accuracy as a function of number of processed world states. Average accuracy of (\textbf{C}) and difference between (\textbf{D}) MetaCOG and Lesioned MetaCOG as a function of faultiness ($\zeta$) in the detections.}
\label{Fig:old_sims_results}
\end{figure*}

The results of Exp. 1 show that MetaCOG can efficiently learn a meta-cognitive representation for an object detector and use it to improve detection accuracy. This provides proof-of-concept that MetaCOG can be applied to neural network object detectors. However, the three networks tested in Exp. 1 do not capture the full space of possible detector performances. How robust is MetaCOG to different degrees of error in its inputs?

To evaluate MetaCOG's robustness, Exp. 2 tests MetaCOG paired with a large range of possible simulated object detectors. To achieve this, we created a synthetic dataset of world states passed through simulated faulty object detectors, producing simulated detections. To reduce computational cost, we used an abstraction of world states and detections (removing all spatial components, detailed below) and used a lightweight, general-purpose version of MetaCOG (see \ref{Lightweight METAGEN}) to learn a meta-cognitive representation of the synthetic detectors and to infer world states.

\paragraph{Dataset} To focus on the contribution of meta-cognition for determining the presence or absence of different object categories in a way that is robust to the failures of the object detector, the dataset consisted of hypothetical collections of objects with no spatial information, passed through simulated object detectors with varying degrees of faulty performance. Specifically, we sequentially sampled world states (vectors of 1s and 0s indicating the presence or absence of five possible object categories), object detectors (miss and hallucination distributions $V$ with a wide variety of parameters $\theta$), and then generated faulty detections (vectors of 1s and 0s indicating the detection or lack of detection of each object category) by passing world states through the probabilistic simulated object detectors. The final dataset consisted of 40000 randomly sampled object detectors, each processing 75 world states, with 5-15 processed frames (detection vectors) per world state. The details of synthesizing this dataset are left to \ref{Synthesizing the Dataset for Lightweight METAGEN}.

\paragraph{Comparison models}

As in Exp. 1, we compare MetaCOG to Lesioned MetaCOG, which fixes the parameters $\theta$ of the meta-cognitive representation $V$ to the expectation of the prior over $\theta$. This enables a fair comparison where the two models are matched for complexity. For details on the prior, see \ref{Lightweight METAGEN}, and for model details, see  \ref{Comparison models for Lightweight METAGEN}.

\paragraph{Results}
Here, we demonstrate that, as in Exp. 1, MetaCOG can infer a detector's hallucination and miss rates without feedback, and that its improvements in accuracy track with its learning of a meta-cognitive representation. See \ref{Metrics for Exp 2} for definitions of the metrics used.

Fig. \ref{Fig:old_sims_results} shows MetaCOG's performance over the 40000 sampled object detectors. To assess MetaCOG's ability to learn an accurate meta-cognitive representation $V|\theta$, we examined the MSE of $\hat{\theta}$ as a function of the number of observed videos. Fig. \ref{Fig:old_sims_results}A shows that MetaCOG's estimates $\hat{\theta}_t$ of the hallucination and miss rates rapidly approach the true values $\theta$, with a final average MSE of 0.0017. To test if learning varied as a function of the object detector's errors, we next calculated MSE as a function of the detector's faultiness $\zeta$, defined as the detector's average proportion of errors per scene (see \ref{Metrics for Exp 2} for formal definition). A linear regression predicting MSE as a function of faultiness revealed that MetaCOG's learned $\hat{\theta}$ is less accurate for faultier detectors ($\beta=0.002$; $p<0.001$), although the effect was minimal (predicting a MSE increase of $0.001$ from a perfect detector with faultiness 0 to a detector with faultiness 0.5, where detections are equally likely to be true or false). Together, these analyses confirm that the results of Exp. 1 hold for a wider set of object detector performances.

Fig. \ref{Fig:old_sims_results}B shows the average accuracy of the models' inferences about world states. The yellow line shows MetaCOG's rapid increase in accuracy over the first 40 world states. Over the course of these 40 world states, MetaCOG's accuracy increases from 88.8\% to 93.6\% (with 93.8\% final accuracy after the 75th world states). This increase in accuracy occurs simultaneously with the decline in MSE for the inferred parameters of the meta-cognitive representation. By comparison, the blue line shows Lesioned MetaCOG's performance, which showed an average performance of 91.2\%. The green line shows MetaCOG's accuracy after conditioning on the parameters $\hat{\theta}_T$ of meta-cognitive representation learned after observing the detections from all $T$ world states (see \ref{Comparison models for Lightweight METAGEN} for details).

To quantify MetaCOG's robustness to faulty inputs, we next computed accuracy as a function of the object detector's faultiness $\zeta$ on a range from 0 (perfect performance) to 0.5 (detections equally likely to be true or false). Figure \ref{Fig:old_sims_results}C shows MetaCOG's and Lesioned MetaCOG's mean accuracies as a function of $\zeta$. When faultiness is low, MetaCOG and Lesioned MetaCOG perform near ceiling, as the object detector's output is already highly reliable. However, MetaCOG outperforms Lesioned MetaCOG in accuracy for detectors with faultinesses in $\zeta \in [0.12,0.5]$.

Fig. \ref{Fig:old_sims_results}D shows the average difference in accuracy between MetaCOG and Lesioned MetaCOG as a function of faultiness. MetaCOG reaches its highest accuracy boost over Lesioned MetaCOG at faultiness level 0.42 (with a 21.8\% improvement), and consistently shows a performance boost across a wide range of faultiness values. Together, these results show that MetaCOG's success  is not limited to the particular neural networks tested in Exp. 1. Instead, MetaCOG can learn a good meta-cognitive representation for a wide range of detectors and use it to improve accuracy. 
\section{Discussion}\label{discussion}

In order to behave intelligently in a complex and dynamic world, autonomous systems must be able to account for inevitable errors in perceptual processing. In humans, meta-cognition provides this critical ability. Inspired by human cognition, we formalize meta-cognition and apply it to the domain of object detection. Our work is a proof-of-concept of how a meta-cognitive representation can improve the accuracy of object detection systems. The MetaCOG model presented here can be directly applied to embodied systems, and MetaCOG may be particularly useful in situations where the vision system may be unreliable, like if the agent is deployed in an environment that was not adequately represented in training. Outside of this specific use-case, the more general MetaCOG approach can support robust AI more broadly.


First, the MetaCOG approach provides a way to quantify uncertainty, even for a pre-trained system and when ground-truth is not accessible. Sudden changes in the uncertainty expressed in the meta-cognitive representation could be used to detect domain or distribution shifts, so as to flag situations of high uncertainty that pose increased risk of unreliable behavior. This could be useful for downstream decision-making processes determining when to halt action.

Second, the MetaCOG approach can be used to tune and improve the underlying system that it is monitoring, as when MetaCOG's inferences were used for fine-tuning Faster R-CNN (Fig. 3). Using meta-cognition to generate a training signal is a novel alternative to human-annotation. In domains where the ground-truth labels required for traditional training are difficult to acquire, the MetaCOG approach could prove especially valuable.

Finally, our work focused on learning a meta-cognition for the purpose of improving accuracy and robustness, but the meta-cognitive representation also has potential benefits for transparency and interpretability. In our implementation, MetaCOG's $V$ captures the object detector's performance in a way that is easy for humans to understand. As such, this approach of learning a meta-cognitive representation may be able to support explainable AI by generating simplified meta-representations of a black-box system's performance.

\section{Conclusion}

We proposed a formalization of meta-cognition for object detectors that increases accuracy by removing hallucinations and filling in missed objects. Our model, MetaCOG, learns a probabilistic relation between detections and the objects causing them (thereby representing the detector's performance and instantiating a form of \textit{meta-cognition}). This is achieved as joint inference over the objects and a meta-cognitive representation of the detector's tendency to error. Critically, MetaCOG performs this inference without feedback or ground-truth object labels, instead using cognitively-inspired priors about objects. Applying MetaCOG to three neural object detectors (Exp. 1) and to simulated detectors (Exp. 2) showed that MetaCOG can efficiently learn an accurate meta-cognitive representation for a wide range of detectors and use it to account for errors, correctly inferring the objects in a scene in a way that is robust to the faultiness of the detector. This work is a proof-of-concept that meta-cognition can be used to quantify and correct a system's faults in the domain of computer vision and perhaps even beyond.


\begin{contributions} 
    MDB developed the methodology, wrote the model code, conducted the experiments and analyses, and wrote the paper. ZA contributed to the code and methodology. MB provided technical and methodological advice (including the Random Finite Sets package in Julia) and edited the paper. ZT suggested experiments and helped with framing and writing the paper. JJE conceived the idea, supervised the project, and wrote the paper.
\end{contributions}

\begin{acknowledgements} 
    This work was supported by NSF award IIS-2106690. We thank the Yale Center for Research Computing, specifically Tom Langford and Kaylea Nelson, for guidance and assistance in computation run on the Milgram cluster. Thanks to Aalap Shah, Eivinas Butkus, and Ilker Yildirm for helpful conversation and comments. We'd also like to acknowledge research assistants Ben Sterling, Bernardo Eilert Trevisan, and Tanushree Burman. Finally, thanks to Vatsal Patel for lending his sharp geometric reasoning!
\end{acknowledgements}

\bibliography{refs}

\newpage

\onecolumn

\title{MetaCOG: A Hierarchical Probabilistic Model for Learning Meta-Cognitive Visual Representations\\(Supplementary Material)}
\maketitle

\appendix

\vspace{1.0pt}
\section{Additional MetaCOG Details} \label{Gen Julia}

The MetaCOG model and inference were implemented in the Julia-based probabilistic programming language Gen \citep{Cusumano-Towner:2019:GGP:3314221.3314642}, licensed under the Apache License Version 2.0. Code and data are available in our repository:  \url{https://github.com/marleneberke/MetaCOG/tree/main}

\subsection{Details of Procedure for Taking the Difference Between $W_t$ and $D_t$} \label{diff between Wt and Ot}
From a world state $W_t$ and a corresponding collection of detections $D_t$, MetaCOG tries to infer which detection were hallucinations or misses so as to update the prior over the hallucination and miss rate. The procedure for this inference is as follows: first, the objects in $W_t$ are projected from 3D space to the 2D image. Then, each detection is matched to the nearest projected object of the same category as long as the Euclidean distance between the pair is within radius $\sigma = 200$ pixels (equal to the spatial noise parameter). After matching, the number of unmatched objects per category are counted, and considered to have been missed. The number of unmatched detections are also counted, and considered to be hallucinations. This process informs the updating of the $\alpha$s and $\beta$s in the priors described above.

\subsection{Object Constraints and Generative Model Parameters}\label{Additional generative model parameters}

Implementation of object constraints were as follows. World states without object persistence were not included in $\mathbb{W}$, which is equivalent to implicitly setting their prior probability to 0. The assumption that two objects cannot occupy the same region in space was implemented through a prior over $\mathbb{W}$ where the probability of having one object near another decreased to 0 according to a Gaussian distribution  with $\sigma^2 = 1$.

To account for noise in a detector's location detection $D_t$, each detection was modeled as having 2D spatial noise, following a Gaussian with $\sigma_{x,y} = 200$ pixels.

Finally, the full generative model requires specifying a prior distribution over camera positions and focal points (although these are observable), set as uniform over 3D space, and a prior distribution over expected number of objects in a scene, sampled from a geometric distribution with parameter $p = 0.9$ (with a uniform prior over category type).

\subsection{Inference Procedure Implementation}
\label{Inference Procedure Details}

We sequentially approximate the joint posterior given in \ref{sc:inference} using a particle filter with 100 particles. The solution space to the inference problem that we consider is sparse, and Sequential Monte-Carlo methods can suffer from degeneracy and loss of diversity in these situations. Our inference approach solves these problems by implementing particle rejuvenation over objects, locations, and beliefs about $V$. Rejuvenation is conducted using a series of Metropolis-Hastings \citep{canini2009online} moves with data-driven proposals designed to obtain samples from $Pr(\vec{V},\vec{W}|\vec{O}, \vec{cj})$.

During object rejuvenation, a new state $W'_t$ is proposed by either adding or removing an object from the current state $W_t$. The probability $p_{add}$ of proposing to add an object depends on the belief about $V_t$ and the observed detections. The number of detections across the five categories are summed. Call this sum $k$. The probability $p$ of $k$ or more detection being produced by hallucinations is calculated as $p = 1 - e^{-\lambda_{total}} \sum_{i=0}^{k} \frac{\lambda_{total}^i}{i!}$, where $\lambda_{total} = \sum_c(\lambda_c)$. (That $e^{-\lambda_{total}} \sum_{i=0}^{k} \frac{\lambda_{total}^i}{i!}$ term is the cdf of $Pois(\lambda_{total})$ from $0$ to $k$.) $p_{add}$ should depend on $p$ such that, when the belief about the overall hallucination rate is high relative to the number of detections, the probability $p_{add}$ of proposing adding a new object to the world state is low, but when the hallucination rate is low relative to the number of detections, the probability $p_{add}$ of proposing adding a new objects is high. However, if $p_{add}$ were to be set to $p$, then there is a danger that, when there are more detections than can be explained as hallucinations, the proposal function would only ever propose adding objects, and never remove them. To balance the relative proposals of adding and removing objects, we set $p_{add} = 0.5*p$, effectively setting the maximum of $p_{add}$ to 0.5.

Objects in $W_t$ have a weighted probability of being removed to produce $W'_t$, such that their probability of removal is inversely proportional to the number of times the object was observed in scene $t$. New objects are added based on a data-driven distribution. This distribution samples object categories from a categorical distribution biased towards categories observed in the current scene $t$ (see Data-driven proposal functions below). With probability 0.5, location is sampled from a 3-D uniform distribution, and otherwise sampled from a data-driven function with location biased toward 3D points likely to  have caused the 2D detections. The proposal for adding a new object or removing and existing object from the world state is accepted or rejected according to the MH algorithm.

After a proposed change to $W$, a second rejuvenation step is performed on locations, wherein an object in $W$ is randomly selected (with equal probability) to have a new location proposed. With probability 0.5, the new location is drawn from a multivariate normal distribution centered on the previous location, and it is otherwise sampled from a data-driven proposal. The proposed world-state $W'$ with a perturbed location is then accepted or rejected according to the MH algorithm.

Finally, new parameters $\theta'$ for the meta-cognition are proposed by perturbing $\theta$. Each parameter in the pair of length $|C|$ vectors is perturbed, with new values generated for $\theta(i,j)_t$ and for $\theta(i,j)_{t-1}$ $\forall i,j$, where $i$ indexes the two vectors and $j$ indexes their elements. The new value for each $\theta(i,j)_t$ is sampled from the appropriate distribution (Beta for the miss rate, Gamma for hallucination rate) with $\alpha_t$ and $\beta_t$, while the new value for $V(i,j)_{t-1}$ is sampled with $\alpha_{t-1}$ and $\beta_{t-1}$. These new values are accepted or rejected according to the MH algorithm.

This three-step rejuvenation process for world states, locations, and parameters of the meta-cognition is done for each particle for 200 iterations and the last state reached in the chain is used as the new rejuvenated particle.

\subsection{Data-Driven Proposal Functions}\label{Data-Driven Proposal}

During rejuvenation, new objects are proposed to be added to $W_t'$ using a data-driven distribution. The category of the object comes from a categorical distribution. A category's weight in the categorical distribution depends on the number detections of that category in the current scene, and on the belief about the probability that those detections reflect a real object (as opposed to a hallucination). The meta-cognition can be used to calculate belief about the probability that this detection was generated by a real object. By Bayes rule, $P(real|obs) = P(obs|real)*P(real)/P(obs)$. Under the prior, each category has the same probability of appearing, so $P(real)$ will not affect the relative weighting of the categories. $P(obs|real)$ is simply the probability of detecting an object when it's present, $p_c$, for that category. $P(obs)$ is the probability of detecting an object of category c. Since a detection must be either true or a hallucination, $P(obs) = P(obs|real) + P(obs|not \; real) = p_c + (1-e^{-\lambda_c})$, where the $(1-e^{-\lambda_c})$ term is the probability of hallucinating one or more objects of category $c$. All together, the probability of sampling category $c$ is proportional to $\frac{k_c*p_c}{p_c + (1-e^{-\lambda_c})}$, where $k_c$ is the number of detections of category $c$ in this scene. So that its possible (though unlikely) to sample an object from an unseen category, $k_c$ is set to have a minimum of one. This probability distribution for proposing object categories incoorporates both the number of times a category is detected, and the meta-cognitive belief about whether those detections were a mere hallucinations or generated from a real object.

In addition to a category, a location must be sampled for the proposed object. With probability 0.5, the location is sampled from a 3-D uniform distribution over the whole scene, and otherwise sampled from a data-driven function. Using the data-driven function, a point is sampled based on proximity to the line-segment that, when projected onto the 2D image, would result in the point where the detection was observed. The probability of proposing a particular point decreases with the distance from this line segment, following a Gaussian with $\sigma^2 = 0.01$. 

In the second rejuvenation step a new location is proposed for an object. With probability 0.5, the new location is drawn from a multivariate normal distribution centered on the previous location, with $\sigma_{x,y,z} = 0.01$. Otherwise, it is sampled as described in the previous paragraph (based on proximity to the line segment that would results in the detection's 2D location).

\section{Experiment 1 Additional Details}

\subsection{Experiment 1 Dataset} \label{Experiment 1 Dataset}

Evaluating Exp. 1 we needed a dataset of videos of the same COCO object categories (so pre-trained NNs could detect them) across multiple videos. These videos had to include meta-data about the camera trajectory and position. Furthermore, so as to enable use to assess MetaCOG's accuracy locating objects in 3D space, we also needed ground-truth labels for the objects, including their locations in 3D space. Because we did not find any existing datasets to satisfy all these constraints, we rendered our own custom dataset.

\subsubsection{TDW Scene Rendering}\label{Rendering scenes}

The TDW environment is licensed under the BSD 2-Clause ``Simplified" License, Copyright 2020 Massachusetts Institute of Technology. We used the ``box room 2018" model with a footprint with dimensions $\approx 12$ x $8$. For scale, the largest object model is about 1 unit along any face.

The number of objects in the scene is uniformly drawn from the counting numbers up to 3 and each object is uniformly assigned one of 5 object categories: potted plant, chair, bowl, tv, or umbrella. We sequentially place objects at locations drawn from a uniform distribution, resampling if two objects are within 1 units of each other or the walls so as to avoid collisions. 
 
\subsubsection{Camera Trajectory Sampling}\label{camera_trajectory}

A camera trajectory consists of a sequence of camera positions and focal points. We generate frames by querying the camera for an image at 20 linearly-spaced times.

\begin{figure}
\centering
\includegraphics[width=0.48\textwidth]{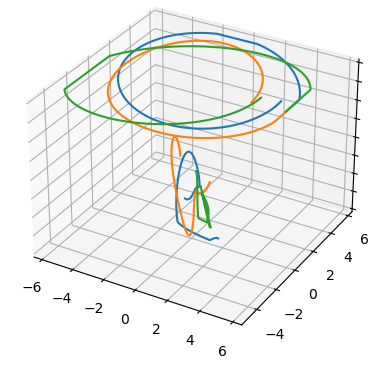}
\caption{Three sampled trajectories. The wide circular patterns at the top are camera positions, and the smaller patterns near the bottom are camera focal points.}\label{Fig:Trajectories}
\end{figure}

Camera trajectories consisted of a circular path around the periphery of the room with noise, generated by a Gaussian process with a radial basis function (RBF) kernel with parameters $\sigma=0.7$ and $\ell=2.5$. The height of the camera is held constant at $y=2$. The focal point trajectory is also sampled from a Gaussian process with an RBF kernel, mean above the center of the floor and component-wise parameters $\sigma=0.7$ and $\ell=2$. Fig. \ref{Fig:Trajectories} shows some example trajectories.

\begin{figure}
\centering
\includegraphics[width=\textwidth]{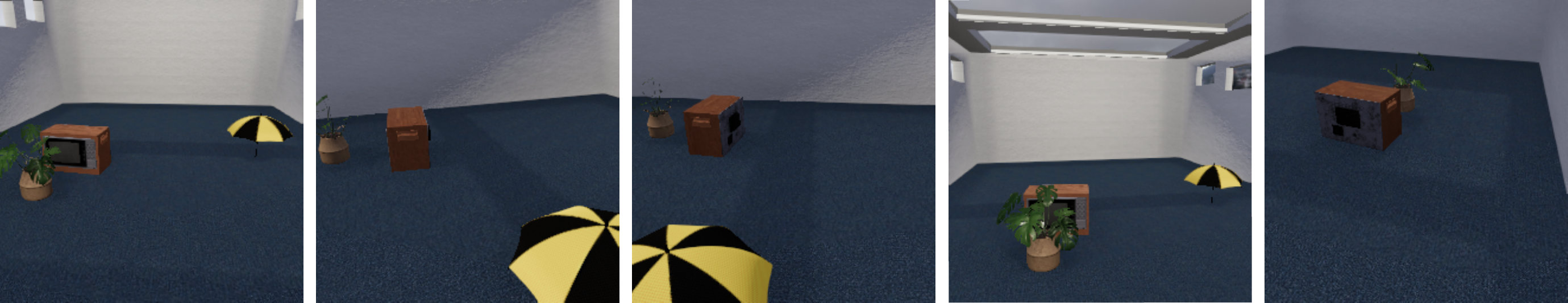}
\caption{Five images from a scene in the dataset. The scene has three objects -- a potted plant, a TV, and an umbrella. In the fourth image, the plant partly occludes the TV, and in the last image, the umbrella is out of view. The rest of the dataset and demo videos can be found at: \url{https://github.com/marleneberke/MetaCOG/tree/main}}
\label{Fig:Images}
\end{figure}

The resulting images include objects passing in and out of frame and sometimes partly occluding each other. Fig. \ref{Fig:Images} shows five images from a scene in the dataset.

\subsubsection{Video Counter-Balancing}\label{appendix:counterbalancingvideo}

To avoid order effects, all tests were run with four different videos orders. Videos were first randomly labeled from 1 to 50, and the four counterbalanced orders were: \{1 ... 50\}, \{50 ... 1\}, \{26 ... 50, 1 ... 25\}, and \{25 ... 1, 26 ... 50\}.

\subsection{Details About Neural Networks Used in Experiment 1}
\subsubsection{Pre-trained weights} \label{pre-trained weights}

For Faster R-CNN, we used the resnet50 pre-trained weights. For RetinaNet, we used the resnet50 pre-trained weights. And for DETR, we used the resnet101 pre-trained weights. These NNs were pre-trained on the COCO dataset.

Faster R-CNN is licensed under The MIT License (MIT), RetinaNet under Apache License 2.0, and DETR under Apache License 2.0.
 
 \subsubsection{Post-Processing} \label{post-processing}
 
 We process the outputs of the object detectors by filtering for the 5 object categories present in the scenes and then performed Non-Maximum Suppression (NMS) with an IoU (Intersection over Union) threshold of 0.4 (applied only for RetinaNet and Faster R-CNN, as it is not typically used for DETR). As input to MetaCOG, we took the top five highest-confidence detections per frame. The reason for taking the top five highest-confidence detection per frame, rather than all of the detections per frame is that sometimes some networks, especially DETR, can output many detections for a single frame. Typically, a confidence threshold is selected by fitting to ground-truth labels. But our setting is unsupervised and meant to apply to situations where ground-truth labels are perhaps unavailable. So, instead of fitting the confidence threshold using ground-truth, we took the top five highest-confidence detections. On most frames and with most networks, the NNs rarely output more than 5 detections, so in practice, taking the top 5 detections rarely affected the NN's output. DETR is the exception, and output more than 5 detections on many frames, resulting in its relatively poor baseline performance, as can be seen in Table \ref{MainTable}.
 
\subsubsection{mAP of NNs on our Dataset} \label{mAP}

To demonstrate that the neural networks are performing reasonably on our dataset, we calculated mAP values for the neural networks when applied to our dataset. We find that the object detectors perform similarly on our dataset as on the datasets on which they are typically evaluated. To show this, we evaluated the object detectors using the standard COCO evaluator (\citeauthor{lin2014microsoft} \citeyear{lin2014microsoft}). On the dataset used in Exp. 1, Faster R-CNN has a mAP of 39.4 on the training set, and 41.1 on the test set. This is similar to the mAPs reported for the most challenging object categories. For instance, Faster R-CNN's reported mAP for the plant category is 39.1 and 40.1 on the PASCAL VOC 2007 and 2012 test sets, respectively (\citeauthor{ren2015} \citeyear{ren2015}). RetinaNet had a mAP of 35.9 on the training set, and 36.3 on the test set. RetinaNet’s reported mAP (across all object categories) on the COCO test dataset is 39.1 (\citeauthor{lin2017} \citeyear{lin2017}), so RetinaNet performs similarly on our dataset as on COCO. DETR had a mAP of 
0.478 on the training set, and 
0.492 on the test set. DETR’s reported mAP on the COCO dataset is 43.5 (\citeauthor{carion2020} \citeyear{carion2020}). This confirms that the object detectors are operating within their expected performance ranges.

 
\subsection{Metrics for Experiment 1} \label{Metrics for Expt 1}

We evaluate MetaCOG's performance in three ways: 1) by testing whether it can learn an accurate meta-cognition of the object detection module, 2) by testing whether the learned meta-cognition improves inferences about which objects are where in 3D space, and 3) by testing whether the learned meta-cognition leads to improved accuracy at detecting objects in the 2D images.

\subsubsection{MSE for $\theta$}\label{MSE for V}

To measure how well MetaCOG learned a meta-cognition, we calculated the MSE between the inferred parameters $\hat{\theta}$ of the meta-cognition and the true parameters $\theta$. The true $\theta$ was calculated for each object detection system as described in the following section \ref{ground-truth V}. The squared error was calculated for corresponding values in $\hat{\theta}$ and the true $\theta$, then summed and divided by the total number of values ($2*|C|$).

\subsubsection{Calculating Ground-Truth $\theta$}\label{ground-truth V}
We needed a way, for a given image, to calculate how many times an object of each category was missed or hallucinated by the object detector. There are many possible ways to define and calculate ground-truth misses and hallucinations, but we chose a definition and procedure to match the definition and procedure MetaCOG uses (see \ref{diff between Wt and Ot}).


We used the 3D position of the object, the camera, and the camera's focus to determine whether and where the object would appear in the image. We considered an object to be missed if the object was in the image, but not detected within 200 pixels of its projected 2D location on the image. We calculated the miss rate for a category $c$ as the number of misses divided by the number of times an object of category $c$ was in view.

By hallucination, we mean a detection of an object of a particular category that was not caused by a ground-truth object of that category. To operationalize this, we considered a detection to be a hallucination if the detection occurred more than 200 pixels away from the projection of a ground-truth object of that category. We calculated the hallucination rate for a category $c$ by counting the number of times an object of category $c$ was hallucinated in our dataset, and then dividing by the total number of frames in the dataset.

The 200 pixel radius was set to match the radius used by MetaCOG in matching $W_t$s to $D_t$s (see \ref{diff between Wt and Ot}).


\subsubsection{Jaccard Similarity}\label{Jaccard Similarity}
The Jaccard similarity coefficient is a useful metric for evaluating inferences about \textit{which} objects are in a frame or scene. The Jaccard similarity coefficients measures the similarity between two sets by taking the size of the intersection of two sets and dividing it by the size of the union of the two sets. We can apply this measure to our setting by treating the object labels that were inferred as one set and the objects that were actually present as the second set. The Jaccard similarity coefficient is given by $
J(\mathbb{I},\mathbb{G}) = \frac{|\mathbb{I} \cap \mathbb{G}|}{|\mathbb{I} \cup \mathbb{G}|}$,
where, in our setting, $\mathbb{I}$ is the set of object labels that were inferred (i.e. $\mathbb{I}$ = \{chair, chair, bowl\}) and $\mathbb{G}$ is the set of objects in ground-truth (i.e. $\mathbb{G}$ = \{chair, bowl\}). (In this example, $J(\mathbb{I},\mathbb{G}) = \frac{2}{3}$.) 

This Jaccard Similarity coefficient can be applied either at the frame or scene level, and is used in both contexts in this paper.

\subsubsection{2D Inference Accuracy}\label{world state accuracy}

To assess the accuracy of of the both MetaCOG and the NNs in a comparable way, we wanted an accuracy metric that could incorporate both \textit{which} object categories are present, and \textit{where} they are. The NNs operate over 2D images, not 3D spaces, so we must make this comparison at the level of 2D frames. The standard metric for object detection accuracy on images is mean Average Precision (mAP), which requires both a ground-truth 2D bounding box and an inferred/detected bounding box. Unfortunately, MetaCOG outputs a point in 3D space, which can be projected to a point in 2D space, but cannot be used to make a bounding box. But we still needed a metric by which to compare MetaCOG's object detection accuracy to those of the neural networks without a meta-cognition. 

To solve this problem, we projected MetaCOG's inferences about the 3D location of objects to a point on each 2D image, and for the neural networks, we took the centroid of the bounding boxes. This way, MetaCOG and the neural networks' detections have the same format: an object label and a point on the image, or a tuple of $(c, x, y)$. Then, for each object category, we used the Hungarian algorithm with Euclidean distances as cost to pair up the detections with the centroid of the ground-truth bounding box. For each pairing, we simply coded whether or not the detected point was within the ground-truth bounding box as $0$ or $1$. Last, we calculated the Jaccard similarity coefficient (see \ref{Jaccard Similarity}) defining pairs as within the intersection of the sets if and only if they pair received the $1$ coding. In other words, we counting up the number of pairs where the detection lay within the ground-truth bounding box, and divided by and the union of the detection and ground-truth sets: the sum of the number of unpaired detections, unpaired ground-truth bounding boxes, and number of total pairings. To extrapolate this accuracy metric from frames to videos, we simply average across frame accuracy for each frame in the video, and we average across videos to calculate overall accuracy. This accuracy metric encodes, for 2D frames, accuracy at detecting both \textit{which} objects are present and \textit{where} they are. 

It is important to note that assessing accuracy at the 2D frame level is different from assessing accuracy at the 3D scene level. Accuracy at the 2D level should depend on the objects that are in the frame. For example, it could be the case that the scene contains a chair and a plant, but the chair is only in frame once, while the plant is in frame 20 times. In that case, 2D accuracy should depend more on identifying and locating the plant than the chair. This distinction will be important when we assess accuracy at the level of the 3D scene in the next section.

\subsection{Experiment 1 Supplemental Results and Details}

\subsubsection{3D Inference Accuracy}\label{3D results}

\begin{table*}[b]
\begin{center}
\begin{small}
\caption{Scene-Level Accuracy Results for Exp. 1. Values in Parentheses Represent 95\% Bootstrapped Confidence Intervals.}
\label{threeD results table}
\begin{sc}
\begin{tabular}{llcc}
\toprule
Object Detector & Model            & Acc. Training                   & Acc. Test\\
\midrule
RetinaNet       & MetaCOG          & \textbf{0.66} (0.58,  0.75)  & \textbf{0.60} (0.51, 0.68)\\
                & Lesioned MetaCOG & 0.58 (0.49, 0.66) & 0.48 (0.41, 0.55) \\
Faster R-CNN     & MetaCOG & \textbf{0.76} (0.67, 0.84) & \textbf{0.66} (0.58, 0.74)\\
                & Lesioned MetaCOG & 0.58 (0.49, 0.66) & 0.49 (0.43, 0.56)\\ 
                
DETR            & MetaCOG          & 0.32 (0.26, 0.38) & \textbf{0.35} (0.29, 0.41) \\
                & Lesioned MetaCOG & \textbf{0.38} (0.30, 0.46) & 0.34 (0.28, 0.41) \\
\bottomrule
\end{tabular}
\end{sc}
\end{small}
\end{center}
\end{table*}

Our main results focused on MetaCOG's accuracy in 2D images. This was necessary so that we could evaluate its inferences against \textit{NN Output}, because the object detection neural networks that we use only provide location on 2D images. Although not our main focus, here we report supplemental results evaluating MetaCOG's capacity to infer 3D world states in Exp. 1. This allows us to test MetaCOG's inferences at the scene level (i.e., what objects are in the room and where are they?), rather than at the frame level (i.e., what objects are visible at this time point and where are they in this 2D projection?). Throughout, we compare MetaCOG only to Lesioned MetaCOG, as it is the only comparison model that also produces results in 3D space.

We first evaluate MetaCOG's accuracy in determining \textit{which} objects are present in a scene (independent of location). To do this, we calculated the Jaccard similarity coefficient (see \ref{Jaccard Similarity}) between the object categories inferred to be present in the scene, and those actually present. Table \ref{threeD results table} shows Jaccard similarity coefficient averaged across the scenes. We find that on the whole, MetaCOG outperforms Lesioned MetaCOG, except for when paired with DETR, where the two perform comparably (as can be observed by noting that the confidence interval of each model's accuracy contains the mean of the other, indicating that there is no significant different between the means).

Next, we evaluate MetaCOG's accuracy in determining \textit{where} the objects are in the scene. To do this, we first paired the inferred and ground-truth object locations of the same category using the Hungarian algorithm with Euclidean distance as the cost. We then averaged the distance between all of the pairs, in both the training and test sets, resulting in an average distance between inferred and ground-truth objects. The results in are shown in Table \ref{distances_3D}, in the Avg. Distance All Pairs column.

\begin{table*}[t]
\begin{center}
\begin{small}
\begin{sc}
\caption{Average 3D Distance Between Inferred and Ground-Truth Object Locations.}
\label{distances_3D}
\begin{tabular}{llcc}
\toprule
Object Detector & Model            & Avg. Distance All Pairs       & Avg. Distance Shared Pairs\\
\midrule
RetinaNet       & MetaCOG          & \textbf{0.37} (0.31, 0.42) & \textbf{0.30} (0.26, 0.34) \\
                & Lesioned MetaCOG & 0.58 (0.49, 0.66) & 0.48 (0.41, 0.55) \\
Faster R-CNN    & MetaCOG & 0.36 (0.30, 0.40) & \textbf{0.30} (0.25, 0.34) \\
                & Lesioned MetaCOG & \textbf{0.31} (0.25, 0.35) & 0.31 (0.25, 0.35) \\       
DETR            & MetaCOG          & 1.26 (0.90, 1.58) & \textbf{0.71} (0.45, 0.92) \\
                & Lesioned MetaCOG & \textbf{0.83} (0.52, 1.11) & 0.81 (0.51, 1.06) \\
\bottomrule
\end{tabular}
\end{sc}
\end{small}
\end{center}
\end{table*}

On the whole, Lesioned MetaCOG's meta-cognition represents a system with a relatively high hallucination rate (higher than the rates inferred by MetaCOG). That belief leads Lesioned MetaCOG to only infer that an object was present when there were more detections of that object, and more spatial information, leading to greater location precision for the objects it inferred to be present. Consequently, Lesioned MetaCOG only infers the presence of the objects that are easiest to locate.

For a more fair comparison, we can examine the average distance between pairs of inferred and ground-truth objects only for ground-truth objects that both MetaCOG and Lesioned MetaCOG's inferred to be present. These average distances are reported in the Avg. Distance Shared Pairs column of Table \ref{distances_3D}. For objects that both MetaCOG and Lesioned MetaCOG inferred to be present, MetaCOG consistently infers slightly better locations than does the Lesioned model.

For reference, the footprint of the room is $\approx12$ x $8$, and the largest object is 1 unit across.


\subsubsection{Results on Additional Dataset}\label{Additional dataset}

To address questions of how well our results will generalize to a new dataset, we rendered a new dataset in ThreeDWorld, this time using five different background rooms and including up to six objects per scene. This collection of room models varies in room size, texture and color (e.g. carpeting vs wood vs vinyl floors), windows, and even ceiling beams. With the inclusion of more objects and sometimes smaller rooms, this means that the new dataset also includes scenes that are more crowded and complex. As a conservative evaluation we used Faster R-CNN, which has the highest baseline performance of all three neural networks that we use, and for which MetaCOG showed the smallest improvement in the one-room dataset. Table 5 reports the results with Faster R-CNN and MetaCOG on this new dataset. As this dataset is more challenging, Faster R-CNN performs worse on this dataset, so the quality of the inputs to MetaCOG is lower. As the problem is harder, we increased the computation during inference from 200 iterations of the three-step rejuvenation process to 1000 iterations, but no other hyperparameters were changed. 

\begin{table*}[b]
\begin{center}
\begin{small}
\begin{sc}
\caption{Results on Dataset with Multiple Rooms}
\label{Multi-room dataset}
\begin{tabular}{llcc}
\toprule
Model            & Acc. Training                & Acc. Test \\
\midrule
MetaCOG          &\textbf{0.69} (0.64, 0.74) & \textbf{0.60} (0.54, 0.67) \\
Lesioned MetaCOG & 0.49 (0.43,  0.56)        & 0.46 (0.39, 0.53) \\
Faster R-CNN     & 0.66 (0.61, 0.70)         & 0.57 (0.51, 0.63) \\
\bottomrule
\end{tabular}
\end{sc}
\end{small}
\end{center}
\end{table*}

\subsubsection{Faster R-CNN Fine-tuning Details}\label{Training Faster R-CNN}

We trained Faster R-CNN on the training set using MetaCOG's inferences as ground-truth labels. Since MetaCOG infers an object category and a 3D point, rather than a bounding box, we constructed a 50x50 pixel bounding box centered on the MetaCOG inferred's 3D point projected onto the 2D image. Then, Faster R-CNN with initialized to its pre-trained weights was trained on the 1000 images (50 videos $*$ 20 frames) in the training set. The training set was augmented by flipping the images horizontally. Faster R-CNN was trained for 10 epochs with a SGR optimizer (learning rate = 0.005, momentum = 0.9, weight decay = 0.0005) and a learning rate scheduler (step size = 3, gamma = 0.1). 

Faster R-CNN was trained using standard multi-task loss function given in the original Faster R-CNN paper \citep{ren2015}, Eq. 1:

$$L({p_i},{i_i}) = \frac{1}{N_{cls}} \sum_{i}L_{cls}(p_i, p^*_i)  + \lambda \frac{1}{N_{reg}} \sum_{i} p^*_i L_{reg}(t_i,t^*_i)$$

From \citep{ren2015}: ``Here, $i$ is the index of an anchor in a mini-batch and $p_i$ is the predicted probability of anchor $i$ being an object. The ground-truth label $p_i^*$ is 1 if the anchor is positive, and is 0 if the anchor is negative. $t_i$ is a vector representing the 4 parameterized coordinates of the predicted bounding box, and $t_i^*$ is that of the ground-truth box associated with a positive anchor. The classification loss $L_{cls}$ is log loss over two classes (object vs. not object). For the regression loss, we use $L_{reg}(t_i, t_i^*) = R(t_i - t_i^*)$ where R is the robust loss function (smooth L1)... The term $p_i^*  L_{reg}$ means the regression loss is activated only for positive anchors $(p_i^*  = 1)$ and is disabled otherwise $(p_i^* = 0)$. The outputs of the cls and reg layers consist of $p_i$ and $t_i$ respectively." Thus, MetaCOG's inferences are integrated into Faster R-CNN's training by substituting them in for ground-truth in the loss function.

Unlike in traditional fine-tuning regimes, the classification head layer was not replaced because we wished to retain the information about the COCO object categories Faster R-CNN had been pre-trained on, as the object categories in our dataset were a subset of COCO categories.

\begin{table*}[h]
\begin{center}
\begin{small}
\begin{sc}
\caption{Faster R-CNN Fine-tuned on MetaCOG's Inferred Labels.}
\label{retraining_table}
\begin{tabular}{lcccr}
\toprule
Model & Acc. Training & Acc. Test \\
\midrule
Pre-trained NN                & 0.75 (0.67, 0.80)          & 0.69 (0.63, 0.74) \\
MetaCOG                       & 0.79 (0.73, 0.84)          & 0.76 (0.70, 0.82) \\
NN fine-tuned with MetaCOG    & 0.84 (0.81, 0.87)          & 0.81 (0.78, 0.84) \\
MetaCOG Round 2               & \textbf{0.87} (0.83, 0.90) & \textbf{0.85} (0.82, 0.87) \\
\bottomrule
\end{tabular}
\end{sc}
\end{small}
\end{center}
\end{table*}

Recall from \ref{Dataset Exp 1} that all reported results show averages across four runs of MetaCOG, with four different orderings of the videos. Each run of MetaCOG produced slightly different inferences, so we trained Faster R-CNN using the outputs of each of the four runs of MetaCOG, producing four sets of weights for Faster R-CNN. In the third line of Table \ref{retraining_table}, we report the accuracy of fine-tuned Faster R-CNN by averaging across the accuracy of Faster R-CNN with each of the four sets of weights.

Finally, the outputs of fine-tuned Faster R-CNN were input again into MetaCOG, and MetaCOG inferred the objects present. The fourth line of Table \ref{retraining_table} reports the accuracy of MetaCOG, averaged over four runs of MetaCOG, each with input from Faster R-CNN with different weights.

The results are shown in Fig. \ref{Fig:retraining} and Table \ref{retraining_table}.

\subsubsection{Lesioned MetaCOG} \label{lesion}

To whether learning the correct parameters $\theta$ for the meta-cognition actually matter, or if the priors (over $\theta$ and over world states in the form of object constraints) do all the work of improving accuracy, we tested a \textit{lesioned} version of MetaCOG fixing the values in $\theta$ to the mean of the prior. The initial prior over the hallucination rate for each object category is a Gamma distribution with $\alpha = 1$ and $\beta = 1$. The mean of that prior is then $1.0$, so in Lesioned MetaCOG, the hallucination rates are set to $1.0$
The initial prior over the miss rate for each object category is a Beta distribution with $\alpha = 1$ and $\beta = 1$. The mean of that prior is then $0.5$, so in Lesioned MetaCOG, the miss rates are set to $0.5$. So this lesion produces the expectation that an object of each category will be hallucinated on average once per frame, and that, when an object is in view, it has a 50\% chance of being missed.

\subsubsection{Explaining Lesioned MetaCOG's Performance when Paired with DETR} \label{failure}

As described in \ref{lesion}, the Lesioned MetaCOG model's meta-cognitive representation of the behavior of the object detectors is fixed to the mean of the priors, which is a hallucination rate of $1.0$ and a miss rate of $0.5$ for each object category. It just so happens that, because of the post-processing procedure used for DETR (described in \ref{post-processing}), that representation are not too far off the ground-truth. The ground-truth miss rates for DETR are indeed around $0.5$ for most categories (as can be seen in Fig. \ref{Fig:Exp1Results}A), and for one category (umbrella), the hallucination rate is indeed very high $\lambda_{umbrella} = 0.79$. It's worth noting that the full MetaCOG model badly underestimates this category's hallucination rate (estimating it as $\hat{\lambda_{umbrella}} = 0.25$). It seems that DETR's poor baseline performance (Table \ref{MainTable}) made it more difficult for MetaCOG to learn an accurate meta-cognitive representation. Since the Lesioned model's meta-cognition is not bad for DETR, and the full MetaCOG model struggled to learn a correct meta-cognition for DETR, the two models performed comparably when inferring the objects present. This shows that, in the case of DETR, learning the meta-cognition from experience did not result in a performance boost above and beyond MetaCOG's priors over meta-cognition and about the world (i.e. object constraints). In the case of DETR, the priors did all of the work.

\subsubsection{Alignment Between the Statistics of the Dataset and MetaCOG's Meta-cognitive Representation} 
\label{Alignment}

It is possible that the distributions in the MetaCOG's meta-cognitive representation might be overfit to the empirical distribution of detections on the dataset used in Experiment 1, and then inflate MetaCOG’s performance. Here, we address this concern. First, the distributions forming the meta-cognitive representation (Poisson for hallucinations, Geometric for detections/misses) were set based on first-principles, and before the dataset was even rendered. However, an overfit could still happen by coincidence. To address this, for each neural network and each object category, we tested whether the empirical distributions of hallucinations and detections could have come from a Poisson and Geometric distribution (respectively).

To test whether hallucinations followed the Poisson distribution, we tallied the number of hallucinations produced by each neural network on each image our dataset, fit a Poisson distribution to each network’s distribution MLE, and then evaluated the quality of the match through a Chi-Squared goodness-of-fit test. Because the sum of Poisson distributions is also a Poisson distribution, we combined the distributions of all object categories and present one test per neural network. The results show that the distribution of hallucinations does not follow a Poisson distribution for Faster-RCNN ($\chi^2 = 29.5; df = 2; p < 1e-06$), RetinaNet ($\chi^2 = 163.4; df = 4; p < 1e-33$), or DETR ($\chi^2 = 2007.7; df = 5; p < 1e-99$).

To test whether detections followed a Geometric distribution, we likewise tallied the number of times each object in the camera's view was detected, and fitted a Geometric distribution to the detections. Because the sum of geometric distributions is not a geometric distribution, we present Chi-squared goodness-of-fit tests at the category level, for a total of 15 tests (five object categories, for three neural networks). These tests confirm that none of the detections for any NN and object category follow a geometric distribution.

For Faster-RCNN the results for the categories ["chair", “bowl”, "umbrella", "potted plant", "tv"] were [$(\chi^2 = 770.5; df = 3; p < 1e-99), (\chi^2 = 355.2; df = 1; p < 1e-78), (\chi^2 = 887.7; df = 2; p < 1e-99)*, (\chi^2 = 274.4; df = 3; p < 1e-58), (\chi^2 =  44.9; df = 2; p < 1e-09)$]. For RetinaNet the results for the categories ["chair", “bowl”, "umbrella", "potted plant", "tv"] were [$(\chi^2 = 325.9; df = 4; p < 1e-68), (\chi^2 = 290.8; df = 2; p < 1e-63), (\chi^2 = 635.9; df = 4; p < 1e-99), (\chi^2 = 247.5; df = 5; p < 1e-50), (\chi^2 = 70.2; df = 3; p < 1e-14)$]. For DETR the the results for the categories ["chair", “bowl”, "umbrella", "potted plant", "tv"] were [($\chi^2 = 61.0; df = 5; p < 1e-11), (\chi^2 = 221.3; df = 5; p < 1e-45), (\chi^2 = 159.3; df = 5; p < 1e-31), (\chi^2 = 176.0; df = 5; p < 1e-35), (\chi^2 = 611.9; df = 5; p < 1e-99)$]. For all neural networks and object categories (15/15), we were able to reject the null that the detections caused by an object come from a geometric distribution.

Note also that all of the results are significant even after applying a Bonferroni correction for multiple comparisons (significant level of .0027 after correcting for 18 tests).

Taken together, this shows that, far from overfitting to the pattern of detections in the dataset, the chosen distributions chosen do not even fit the pattern of detections observed. This demonstrates that MetaCOG is robust to some level of misspecification of the distributions in the meta-cognition, providing further evidence that MetaCOG can handle some mis-alignment between its priors and the pattern of a NN's detections a particular dataset.

\subsubsection{Compute Resources and Runtime} \label{Compute}

We estimate the compute resources used in producing the final results reported in Exp. 1. Running the MetaCOG (and Lesioned) model four times per each of the three neural networks used 36 CPUs with about 5GB memory per CPU on an internal cluster for approximately 100 hours.

The goal of this work is to show an initial proof-of-concept that meta-cognition can be useful for object detection, and so we were not optimizing for efficiency at this time. That said, it is important to consider the resources that MetaCOG uses.

In Experiment 1, MetaCOG used 100 particle filters and 200 MCMC rejuvenation steps applied to each particle. We had access to 36 CPUs, so we were able to parallelize some of the particle filtering steps.

Under this setup, when paired with Faster-RCNN, MetaCOG took an average of ~50 seconds per world state. There is a fair amount of variance (ranging from 20 to 102 seconds) per world state, owing to the complexity of the observations. To understand this variability, we have to delve deeper into the workings of the MetaCOG.

The most computationally costly piece of MetaCOG’s pipeline is in the inference procedure: specifically, calculating the possible associations between detections and objects in the world. MetaCOG models both the object in the world states and the detections as random finite sets. The bottleneck is in calculating the likelihoods of different possible matchings between detections and objects in the world state. Under the semantics of random finite sets \citep{vo2013labeled}, the conditional distribution over a set of observations (in our context, detections) X given a hypothesis of elements E (in our case, putative objects) is calculated and marginalized over all partitions of E to X where each element in X is associated with an element in E. Since, in our case, the distribution of mass over partitions is sparse (i.e. most of the probability mass is concentrated in a small percentage of partitions), we can employ a random walk approach to substantially reduce runtime without impacting accuracy \citep{lovasz1993random, blum2016foundations} by reducing the complexity from $O(N!)$ to $O(N^2)$.

Aside from the number of detections and putative objects in the world state, MetaCOG’s runtime depends on the amount of compute and the computational resources available. Our method uses a particle filter and MCMC rejuvenation steps applied to each particle. How many rejuvenation steps and particles are necessary for convergence depends on the dataset and the user’s trade-off between speed and accuracy. In Experiment 1, we were optimizing for accuracy in the speed-vs-accuracy trade-off, and so we erred on the side of overkill, using 100 particles and rejuvenating each particle using 200 MCMC steps. A use-case in which less compute is available could set a lower computational budget perhaps without compromising much accuracy. Extending the use-cases to situations in which speed is important requires studying how different computational budgets (i.e. different numbers of particles and MCMC steps) affect the speed-accuracy trade-off.

Finally, MetaCOG may not need to run on every single frame in a video. This approach of thinning frames has been taken to speed up other object detectors to real-time (\url{https://google.github.io/mediapipe/solutions/objectron.html}).

\section{Experiment 2 Additional Details}
\subsection{Lightweight MetaCOG} \label{Lightweight METAGEN}

In addition to the full MetaCOG model described in the main text, we also implemented a simplified version for the simplified setting used in Exp. 2, eliminating spatial information so as to isolate the contribution of meta-cognition. 
This simplified setting is object detection without location, wherein a black-box object detector generates labels (without associated locations) for the objects present in a scene. So detections are simply object labels. Furthermore, the goal of inferences in this simplified setting is not to infer what objects are where in 3D space, but merely which objects are present in the scene. Camera trajectories are no longer included, and all objects are assumed to remain in view at all times.

Additionally, the meta-cognition in this Lightweight MetaCOG is simplified as well: the prior over the meta-cognition, $G$, does not vary with time, but is fixed as a Beta distribution with parameters $(\alpha = 2, \beta = 10)$. Hallucination and misses are treated as Bernoulli random variables, with their parameters $\theta$ sampled from the Beta prior, separately for each object category.

The inference targets and procedure are largely unchanged, with the estimate of joint posterior now sequentially approximated via:

\begingroup
\setlength\abovedisplayskip{0pt}
\setlength\belowdisplayskip{0pt}
\begin{equation*} \label{eq:simple_eq}
Pr\left(\vec{V}, \vec{W} | \vec{D}\right) \approx \prod_{t=1}^{T} Pr\left(D_t | \hat{V_t}^t, \hat{W_t}^{t} \right) Pr\left(\hat{W_t}^t\right) Pr\left({\hat{V_t}}^t | {\hat{V_{t-1}}}^t\right)
\end{equation*}
\endgroup
where  the transition kernel, $Pr(\hat{V}^t | \hat{V}^{t-1})$, defines the identity function.

\subsubsection{Details of Inference Procedure for Lightweight MetaCOG}\label{lightweight_details}

The above equation is estimated via particles filtering, with 100 particles.

We implemented rejuvenation using a series of Metropolis-Hastings MCMC perturbation moves over $\hat{\theta}$. 
The proposal function is defined as a truncated normal distribution with bounds $(0, 1)$:

\begin{equation*} \label{eq:v_proposal}
    \hat{\theta_{i,j}}^{t'}  \sim \mathcal{N}(\mu = \hat{\theta_{i,j}}^{t}, \sigma^2 = 0.01)
\end{equation*}
where $\theta_{i,j}$ is the $j$th element of vector $i$ in the pair of vectors of parameters $\theta$. A proposal is accepted or rejected according to the Metropolis-Hastings algorithm. Each element in $\theta$ is rejuvenated separately and in randomized order.

\subsection{Experiment 2 Dataset}
\label{Synthesizing the Dataset for Lightweight METAGEN}

Here we discuss how we synthesized the dataset for evaluating Lightweight MetaCOG in Exp. 2.

In this context, the object detector can be represented by its parameters, $\theta$, which is a pair of vectors of length $|C|$ containing the hallucination and miss rates for each object category. The parameters for each object detector was generated by drawing ten independent samples from a beta distribution, $\sim \mbox{B}(\alpha = 2, \beta = 10)$.
This distribution allows us to sample object detectors with variable error rates (mean value = $\frac{1}{6}$) while maintaining a low probability of sampling object detectors that produced hallucinations or misses more often than chance (0.005 chance of sampling values above 0.5; 0.06 chance that complete sampled object detector has at least one hallucination or miss rate above 0.5).

For each object detector, we sampled 75 world states. A Poisson distribution $N \sim Poisson(\lambda = 1)$ truncated with bounds $[1, 5]$ determined the number of objects in a world state. The object categories were samples from a uniform distribution. Each world state was a hypothetical collections of objects, summarized as a vector of 1s and 0s indicating the presence or absence of each category of objects. For each world state we used the object detector to synthesize the detections from $5-15$ simulated frames (number sampled from a uniform distribution), producing a total of $375-1125$ simulated frames per object detector. Inferences about the hallucination and miss rates of each object category are independent, and we thus considered situations with only five categories.

\subsection{Metrics for Experiment 2}\label{Metrics for Exp 2}
To measure how well MetaCOG learned a meta-cognition, we calculated the mean squared error (MSE) between the inferred parameters of the object detector $\hat{\theta}$ and the true parameters $\theta$ generating the percepts given by
\begin{equation*}
    \mbox{MSE}=\frac{1}{2|C|}\sum_{c \in C} \bigg( (H_c - \hat{H}_c)^2 + (M_c - \hat{M_c})^2)\bigg)
\end{equation*}
where $C$ is the the set of object classes, $H_c$ is the hallucination rate for category $c$, and $M_c$ is the miss rate for category $c$.

Accuracy was measured using the Jaccard similarity coefficient of the set of object classes in the ground-truth world state $W_t$ and the set of object classes in the inferred world state $\hat{W}_t$, see \ref{Jaccard Similarity}.

To analyze MetaCOG's accuracy as a function of faultiness in the detected labels, we computed the average faultiness in a collection of detections $D_t$ as
\begin{equation}
\zeta_t = \frac{1}{|D_t||C|}\sum_{c \in C}\sum_{x \in D_t}|\mathds{1}_{W_t}(c) - \mathds{1}_x(c)|
\end{equation}
where $C$ is the set of object classes, $D_t$ is the collection of detected labels generated from world state $W_t$, $\mathds{1}_{W_t}(c)$ is an indicator for whether an object of class $c$ is in $W_t$, and $\mathds{1}_x(c)$ is an indicator for whether an object of class $c$ is in the detection $x$.

Because our sampling-based dataset generation process (\ref{Synthesizing the Dataset for Lightweight METAGEN}) does not guarantee enough data points for every possible faultiness value, analysis in Fig. \ref{Fig:old_sims_results}C and D were computed using a rolling window such that each point shows average accuracy on the $[\zeta-.05,\zeta+.05]$ range.

\subsection{Comparison models for Experiment 2}\label{Comparison models for Lightweight METAGEN}
To test whether learning a meta-cognition improves inferences about the world state, we compare MetaCOG with and without the meta-cognitive learning. We call MetaCOG without learned meta-cognition Lesioned MetaCOG. Like the other MetaCOG models, Lesioned MetaCOG has a meta-cognitive representation of $V$ and uses an assumption of object permanence to infer the world states causing the detected labels. Lesioned MetaCOG, however, does not learn or adjust the parameters $\theta$ in its meta-represetation based on the observed labels. Formally, Lesioned MetaCOG assumes that the hallucination and miss rates for every category are the mean of the beta prior over hallucination and miss rates, call it $\hat{\theta}_{0,\mu}$. Lesioned MetaCOG then uses the same particle filtering process described in \ref{eq:simple_eq}, except that it conditions on $\hat{\theta}_{0,\mu}$ instead of $\hat{\theta}_{T,\mu}$.

As described in \ref{Lightweight METAGEN}, the prior over both the hallucination rate and the miss rate is a Beta distribution parameters $(\alpha = 2, \beta = 10)$. The mean of this prior is then $\frac{1}{6}$. So in the Lesioned MetaCOG for Exp. 2, the hallucination rate and the miss rate are both set to $\frac{1}{6}$.

We also compare two variations of MetaCOG with learning. MetaCOG Learning performs a joint inference over $\theta$ and $\vec{W}$ based on collections of detections $\vec{D}$ generated from the sequence of world states $\vec{W}$. This model allows us to evaluate how MetaCOG's inferences improve as a function of the observations it has received. We name this model observation-by-observation inference MetaCOG Learning.

After having received all $T$ observations, MetaCOG could retrospectively re-infer the world states causing the $T$ observations. This model, simply called MetaCOG, re-infers the world states that caused its observations conditioned on its estimate $\hat{\theta}^T_{T,\mu}$, as described in \ref{Estimating V}.

MetaCOG Learning lets us interpret how MetaCOG learns a meta-cognition, and MetaCOG lets us test how MetaCOG performs after having learned that meta-cognition. Lesioned MetaCOG serve as a baseline model for comparison.

\end{document}